\documentclass[journal,10pt]{IEEEtran}
\usepackage{amsthm}
\usepackage{amsmath}
\usepackage{graphicx}
\usepackage{algorithm}
\usepackage{algorithmicx}
\usepackage{algpseudocode}
\usepackage{subfigure}
\usepackage{overpic}
\usepackage{colortbl}

\ifCLASSINFOpdf
\else
\fi
\begin{document}
%
\title{Dynamic Large Language Models on Blockchains}
%
%
%

\author{Yuanhao~Gong\\College of Electronics and Information Engineering, Shenzhen University, China.~~gong.ai@qq.com
	\thanks{Manuscript received April 19, 2005; revised September 17, 2014.}
	}

%
%

\markboth{Journal of \LaTeX\ Class Files,~Vol.~14, No.~8, August~2015}%
{Yuanhao: LLM on Blockchains}
%



\maketitle

\begin{abstract}
Training and deploying the large language models requires a large mount of computational resource because the language models contain billions of parameters and the text has thousands of tokens. Another problem is that the large language models are static. They are fixed after the training process. To tackle these issues, in this paper, we propose to train and deploy the dynamic large language model on blockchains, which have high computation performance and are distributed across a network of computers. A blockchain is a secure, decentralized, and transparent system that allows for the creation of a tamper-proof ledger for transactions without the need for intermediaries. The dynamic large language models can continuously learn from the user input after the training process. Our method provides a new way to develop the large language models and also shed a light on the next generation artificial intelligence systems.
\end{abstract}


%
\IEEEpeerreviewmaketitle

%
%
%
%
\section{Introduction}
The development of artificial intelligence and machine learning has led to the evolution of neural networks. These networks have become more complex due to the growing number of parameters. Usually, larger models can achieve higher accuracy and perform better on complex tasks than smaller models.

The increasing number of parameters in neural networks can lead to challenges in training and deploying these models, as well as difficulties in interpreting the behavior of these increasingly complex models. With the growing number of parameters, neural networks can become computationally expensive, requiring more resources to train effectively. Additionally, the larger size of these models can make them difficult to deploy on devices with limited memory or processing power.


To reduce the computational cost of training and deploying neural networks, researchers are exploring techniques like model compression, which involves reducing the size of the model by removing unnecessary parts, and architecture search, which involves automatically selecting the best architecture for a given task. One approach to reducing the computational cost of training and deploying neural networks is through the use of techniques such as pruning, quantization, and distillation. These methods aim to reduce the number of parameters and operations required by the model while still maintaining its performance. These techniques can help reduce the computational cost of training and deploying neural networks, making them more accessible to a wider range of applications.

The trend towards larger and more complex neural networks presents both challenges and opportunities for the field of artificial intelligence. Researchers are actively working to overcome these challenges and harness the full potential of these models. One way to make neural networks more interpretable is by creating visualizations of the model's internal workings and using techniques like saliency maps to understand how different parts of the input contribute to the output. Additionally, techniques like adversarial training can help improve the robustness of the model and make it more resistant to attacks.

\begin{figure}[!t]
	\centering
	\includegraphics[width=\linewidth]{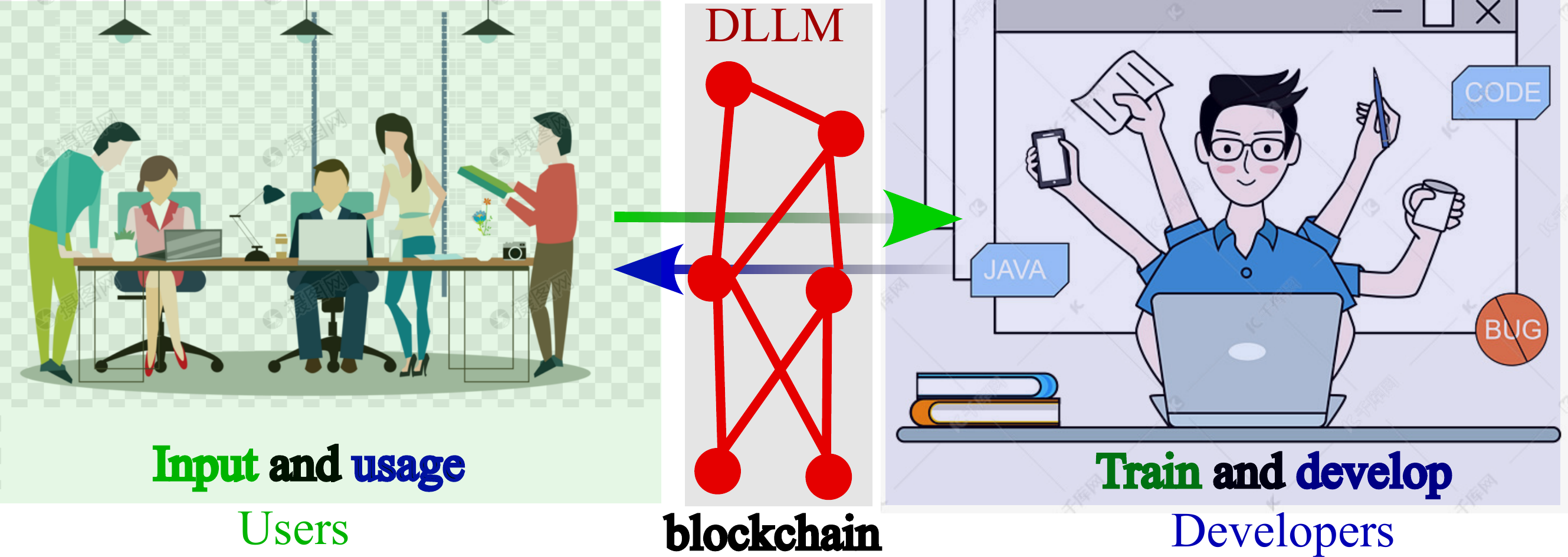}
	\caption{Illustration of the proposed method. The dynamic LLM run on blockchains. Users feed the model more and more data while the developers can update the model accordingly. From economy point of view, users pay for professional advice at a very low price. They earn money for their valuable input. Developers pay for the input to train their models but earn money from their trained models. Such ecosystems will benefit both users and developers.}
\end{figure}

\subsection{Large Language Models}
Large language models (LLMs) are a type of artificial intelligence system that can generate human-like text based on input data. These models are trained on massive amounts of text data and use sophisticated algorithms to learn patterns in language and generate coherent text. LLMs have become increasingly popular in recent years, and their applications range from chatbots and virtual assistants to content creation and language translation.

One of the most significant advantages of LLMs is their ability to generate high-quality text that is indistinguishable from text written by humans. This makes them a valuable tool in industries such as content creation and journalism, where high-quality writing is essential. LLMs can also be used to generate personalized content for consumers, such as product recommendations or personalized news articles.

Another benefit of LLMs is their ability to understand the context of text. These models can analyze the meaning behind words and phrases and generate text that is relevant and accurate. This makes LLMs a valuable tool in industries such as search engine optimization and natural language processing (NLP), where understanding the context of text is critical.

Furthermore, LLMs can be used to automate repetitive tasks, such as customer service inquiries or data entry. This can help businesses save time and money while improving efficiency. LLMs can also be used to generate more accurate translations, making them a valuable tool in the language translation industry.

However, there are some concerns regarding the use of LLMs. One issue is the potential for bias in the data sets used to train these models. Since LLMs are trained on massive amounts of data, any biases in the data can be amplified and reflected in the generated text. Additionally, there are concerns regarding the potential misuse of LLMs to generate fake news or other forms of misinformation.

LLMs are a powerful tool that has the potential to transform various industries by automating tasks, generating high-quality content, and improving efficiency. However, it is essential to address the concerns regarding bias and potential misuse to ensure that these models are used ethically and responsibly.
\subsection{Blockchain}
Blockchain technology has emerged as a groundbreaking concept in the digital world in recent years. It is a secure, decentralized, and transparent system that allows for the creation of a tamper-proof ledger for transactions without the need for intermediaries. This technology holds the potential to revolutionize various industries, such as finance, healthcare, and supply chain management. This essay explores the fundamentals of blockchain technology and its possible impact on different industries.

At its core, a blockchain is a digital ledger of transactions that is distributed across a network of computers. Each block in the chain contains a timestamp and a link to the previous block, forming a chain of blocks. The network of computers that maintains the blockchain ensures that every block added to the chain is valid and unchanged. This makes it nearly impossible to tamper with the ledger or commit fraud, as any change to one block would require a change to every block that follows it.

The potential uses for blockchain technology are vast and varied. In the finance industry, blockchain technology could streamline processes, reduce costs, and increase transparency. Smart contracts, which are self-executing contracts with the terms of the agreement between buyer and seller being directly written into lines of code, could reduce the need for intermediaries and minimize the risk of fraud.

In healthcare, blockchain technology could improve security and privacy. Patient records could be stored on a blockchain, which would allow for secure and easy access to medical records by authorized parties. The technology could also help combat counterfeit drugs by tracking the supply chain and ensuring the authenticity of drugs.

In supply chain management, blockchain technology could improve efficiency and transparency. The technology could be used to track goods from their origin to their final destination, reducing the risk of fraud and ensuring that products are ethically sourced.

Blockchain technology has the potential to transform various industries by increasing efficiency, security, and transparency. While the technology is still in its early stages, many experts believe that it is the future of the digital world. As blockchain technology continues to evolve, it will be interesting to see how it shapes the future of the industries it impacts.
\subsection{Motivation and Contribution}
Traditional LLM is trained on centralized GPUs and can not evolve after the training stage. These drawbacks motivate us to develop LLM on blockchains, where the LLM is trained in decentralized nodes and can evolve during its usage. Our contributions are
\begin{itemize}
	\item  we propose to develop LLM on blockchains
	\item  The proposed LLM can continuously evolve during its usage after the training process
	\item  we propose a new economic model to link the LLM with the blockchain technology.
\end{itemize}

\section{Dynamic Large Language Models}
The development of artificial intelligence has resulted in increasingly complex neural networks with more parameters. This has led to the creation of large language models such as GPT-3 and BERT, which have significantly impacted natural language processing.

However, a major challenge associated with these models is their high computational cost. Traditional large language models, like GPT-3 and BERT, require thousands of GPUs for training and deployment. This high cost is a significant obstacle for many researchers and organizations, limiting access to these powerful models.

Despite this challenge, advancements in hardware and software have made it possible to train and deploy these models on fewer GPUs. This has opened new opportunities for researchers and organizations to explore the capabilities of these models and leverage their potential in various applications.

Another problem with current LLM is that these models are static. In other words, they can not evolve to achieve higher performance with more and more user input. In fact, millions of users can help to improve the LLM in each specific field, such as poem, medical diagnosis and car design.

\subsection{Decentralized Blockchains}
To tackle the heavy computation requirement in LLM's training and deployment, we propose to develop LLM on blockchains which are decentralized and have high computation performance.

Large language models have had a major impact on natural language processing (NLP) in recent years. They've achieved state-of-the-art performance on a wide range of NLP tasks. However, training and deploying LLMs can be challenging due to their massive computational requirements and the need for high-quality training data.

To address these challenges, we propose developing LLMs on blockchains. Blockchains offer a decentralized and transparent platform for storing and processing large amounts of data. By leveraging blockchain technology, we can create a distributed network of nodes that contribute computing power to train and deploy LLMs.

Developing LLMs on blockchains has many benefits. First, it allows for more efficient use of computing resources by leveraging the idle processing power of network participants. Second, it enables the creation of decentralized datasets that are more secure and transparent than traditional centralized datasets. Third, it provides a more fair distribution of economic incentives for data contributors and validators.

One of the main challenges for LLMs is the availability of high-quality training data. With blockchains, we can create a decentralized dataset that is more secure and transparent than traditional datasets. This means that the data is less likely to be tampered with and can be easily audited for accuracy and bias. Also, since the data is decentralized, it is not controlled by any single entity, which makes it more accessible to a wider range of researchers and developers.

Developing LLMs on blockchains also provides a more fair distribution of economic incentives for data contributors and validators. In traditional centralized datasets, data contributors and validators are often not compensated fairly for their contributions. However, with blockchains, data contributors and validators can be rewarded with tokens that have real economic value. This creates a more fair and transparent system that incentivizes participation and collaboration.

Overall, we believe that developing LLMs on blockchains has the potential to revolutionize the field of NLP by making large-scale language models more accessible to everyone. We're excited to explore this new area of research.
\subsection{Dynamic Neural Networks}
To tackle the static issue in LLM, in this paper, we propose dynamic LLM which can evolve after the training process. More specifically, the neural network parameters can be continuously updated during their usage after the training process. In other words, the traditional training process is an initialization of the dynamic LLM. The dynamic LLM have the ability to learn during its usage.

Such behavior is inspired by human brain development which has the life-long learning ability and can increase its knowledge with more and more experience.

Although LLM has much less neurons than the human brain and each neuron in LLM is much simpler than the real neuron in human cortex, LLM can be inspired by the anatomic and functional dynamics of human brain. For example, the fact that human brain can achieve better with more and more experience indicates that LLM should also have the ability to update their parameters with more usage. We call such LLM as dynamic LLM or DLLM for short.

Beside the dynamic parameter weights, DLLM also include a dynamic architecture, which is similar to the functional network on human brain cortex. Such functional network points out that one function is supported by several regions and their connectivity. And the same region might play different roles in different functional networks. Such multi functionalities should be imposed into DLLM for improving the model efficiency and reducing the model size.  
\subsection{Functional Network in Brain Cortex}
The human brain cortex is a highly complex network of functional connections that allow for various cognitive processes, such as perception, attention, memory, and decision-making. Scientists have been able to study these functional connections and identify key brain regions and their connectivity patterns using brain imaging techniques such as functional magnetic resonance imaging (fMRI) and electroencephalography (EEG).

One of the most well-known functional networks in the human brain is the default mode network (DMN). The DMN is active when the brain is at rest and not engaged in any specific task. It includes the medial prefrontal cortex, posterior cingulate cortex, and inferior parietal lobule. This network is believed to be involved in self-referential thinking, mind-wandering, and autobiographical memory.

Another important functional network is the salience network, which is involved in detecting and prioritizing important stimuli. The salience network includes the anterior cingulate cortex and insula. This network is important for our ability to direct our attention towards important stimuli and regulate our emotional responses.

The attention network, on the other hand, is involved in focusing attention on specific stimuli. This network includes regions such as the frontal eye fields and the parietal cortex. The executive control network, on the other hand, is involved in higher-level cognition such as decision-making and working memory. This network includes regions such as the dorsolateral prefrontal cortex and the anterior cingulate cortex.

Functional connectivity studies have also identified other networks in the human brain, such as the language network, the visual network, and the sensorimotor network. These networks are involved in processing language, visual information, and sensory information, respectively. 

These brain networks can inspire to design the modules in LLM and the functions of each module. Although the brain network is far from fully understood, they can guide the structure of LLM and their working fashion, especially when the LLM become larger and more complex.
\section{Training and Running on Blockchains}
After explaining the DLLM in previous sections, we now explain why they should be trained and run on blockchains from several aspects, including transparency, dynamic price, decentralization and economic analysis.

\subsection{Transparency}
Many companies use proprietary data to train their large language models, which can result in highly accurate models. However, this approach also has some drawbacks.

A major issue is that outside researchers often do not have access to the proprietary input used to train these models. This lack of accessibility makes it difficult for others to replicate or independently verify the models' results.

The use of proprietary data can make it challenging to interpret the models. With inputs that are not publicly available, it can be hard to determine the driving factors behind a model's predictions. This lack of transparency can be problematic, particularly when the models make important decisions.

To address these issues, some companies are exploring ways to make their models more transparent and accessible. One method is to fine-tune the models using publicly available data after they have been trained on proprietary data. This can help ensure the models are less reliant on specific inputs and are more generalizable.

In our DLLM, all the input data must be transparent and available to all researchers. Such requirement makes DLLM more transparent than LLM.
\subsection{Dynamic Price}
Unlike previous static LLM, DLLM take dynamic input and are dynamic with different tasks. Therefore, it takes different cost and price for easy and difficult tasks. The more difficult the task, the more computation is required and thus the price is higher for its usage.

Such dynamic price is good for both users and developers. For a easy task such as checking the weather, the user only pays a extremely low price (usually free of charge) because it is not a difficult task. For a more difficult task such as analyzing human's behavior in a set of videos, the user has to pay more money since the task requires more resources.

If the developers build a simple model, they only pay a low price and their model can only earn money at a low price since the model can only deal with simple tasks. In contrast, if they build a complex model, they have to pay a higher price because the complex model requires more computation. As a result, the complex model can earn money at a higher price because it can deal with complex tasks.

Such dynamics also corresponds to the dynamics in the network architecture. Easy tasks only use shallow part of networks while difficult tasks use deep and more networks.
\subsection{Decentralization}
Current LLM are controlled by companies who might block users from using their LLM. And users might lose their private history on the LLM. To tackle this issue, the decentralization nature of blockchains can play an important role here.

The user's private history is encrypted and securely stored using blockchain technology. This ensures that the data is protected from unauthorized access or tampering.

The decentralized nature of the blockchain means that the user's data is not stored in a single location, reducing the risk of data loss due to technical issues or cyber attacks.

The user's data is protected through strong encryption algorithms that make it extremely difficult for anyone to access the information without authorization. The user has full control over their private history and can decide who can view it.

Overall, the use of blockchain technology provides a high level of security and privacy for the user's private history. The user can trust that their data is safe and secure, and they do not have to worry about losing it or being blocked by anyone.

\subsection{Win-win for Everyone}
As mentioned, both users and developers can benefit from DLLM. Users' input and task requirements can be considered as valuable resources for developers. Thus, developers have to pay money for such resources. And the models built from developers are valuable resources. Therefore, the users have to pay money for their usage. 

This is a economical helix structure that benefits everyone in the system. Users get access to the model at a lower price and developers can develop their model at a lower price. Meanwhile, users can use better models if their tasks are more complex. In this case, they have to pay more. If the developers want to build a better model with better input or requiring more computation resource, they also have to pay more.

A better input from users earns more while the better model from developers also earns more. In contrast, the users' input that does not have value for developers will not earn anything and be abandoned. The developers' models that are not required by any users will not earn anything and be inactive. Finally, the system only contains the valuable input from users and models from developers. 

It's important to recognize that both input and models are dynamic, meaning that the popularity and usage of specific models may change over time. Just like fashion trends in human society, machine learning models can come and go.

This is something to consider when developing and implementing models. While a particular model may not be widely used or popular at present, it may become more relevant and widely adopted in the future. Therefore, it's important to continue to explore and evaluate various models, even if they're not currently in use.

By remaining open to new models and ideas, we can ensure that our machine learning systems are up-to-date and effective in a constantly evolving technological landscape.
\section{Conclusion and Discussion}
In this paper, we have presented dynamic large language models that have dynamic architecture and dynamic parameter weights for different tasks. Meanwhile, we propose to perform such models on blockchains, where the decentralization and transparency property can benefit both users and developers.

Our approach sheds light on the next generation of AI systems that are based on blockchain technology. By combining the power of blockchain with AI, we can create systems that are more secure, transparent, and efficient than ever before.

Through our research and development, we have identified important areas where blockchain can be applied to AI systems. These areas include data management, model training and validation, and decision-making processes.

One of the biggest challenges in AI is ensuring the integrity and security of data. By using blockchain, we can create a decentralized and immutable ledger that guarantees the verification and tamper-proof of data. This provides a level of transparency and trust that is crucial for AI systems.

Moreover, blockchain can improve the model training and validation process. By utilizing a distributed ledger, we can guarantee that models are trained on high-quality data and that the results are accurate and reliable.


Our approach is a significant step forward in the development of blockchain-based AI systems. We believe that these systems will revolutionize the way we think about AI and its potential applications in various audio, vision and machine learning tasks \cite{chenouard:2014,gong2009symmetry,Lewis2019,zhao2023survey,Gong2012,Brown2020,gong2013a,Yu2019,Gong:2014a,Yin2019a,gong:phd,Yu2022a,gong:gdp,Guo2022,gong:cf,Zong2021,gong:Bernstein,Ezawa2023,Gong2017a,Tang2021a,Gong2018,Gong2018a,Yu2020,GONG2019329,Sancheti2022,Gong2019a,Tang2021,Gong2019,Yin2019b,Gong2022,Yin2020,Gong2020a,Jin2022,Gong2021,Tang2022,Gong2021a,Tang2022a,Tang2023,Gong2022,Tang2023a,Xu2023,Han2022,Scheurer2023,Zhang2023b}.
\bibliographystyle{IEEEtran}
\bibliography{IEEEabrv,IP}

%




\end{document}